\documentclass[conference]{IEEEtran}
\IEEEoverridecommandlockouts

\usepackage{cite}
\usepackage{amsmath,amssymb,amsfonts}
\usepackage{algorithmic}
\usepackage{graphicx}
\usepackage{textcomp}
\usepackage{xcolor}
\usepackage{texnames}
\usepackage{siunitx}
\usepackage{booktabs}

\usepackage{svg}

\begin{document}

\title{GeoViT: A Versatile Vision Transformer Architecture for Geospatial Image Analysis}

\author{\IEEEauthorblockN{Madhav Khirwar}
\IEEEauthorblockA{
Bangalore, India \\
madhavkhirwar49@gmail.com}
\and
\IEEEauthorblockN{Ankur Narang}
\IEEEauthorblockA{\textit{Founder, Fermion AI}\\
New Delhi, India \\
ankur.narang@fermionai.com}
}

\maketitle

\begin{abstract}
Greenhouse gases (GHGs) are pivotal drivers of climate change, necessitating precise quantification and source identification to foster mitigation strategies. We introduce GeoViT, a compact vision transformer model adept in processing satellite imagery for multimodal segmentation, classification, and regression tasks targeting CO\textsubscript{2} and NO\textsubscript{2} emissions. Leveraging GeoViT, we attain superior accuracy in estimating power generation rates, fuel type, plume coverage for CO\textsubscript{2}, and high-resolution NO\textsubscript{2} concentration mapping, surpassing previous state-of-the-art models while significantly reducing model size. GeoViT demonstrates the efficacy of vision transformer architectures in harnessing satellite-derived data for enhanced GHG emission insights, proving instrumental in advancing climate change monitoring and emission regulation efforts globally.
\end{abstract}


\section{Introduction}
The escalating challenge of climate change is largely fueled by the emission of greenhouse gases (GHGs), primarily from industrial and transportation activities. The criticality of these emissions underscores the need for accurate and efficient monitoring techniques to better understand and mitigate their impact. While deep learning models have shown promise in estimating GHG emissions using satellite imagery, there exists a growing demand for models that are not only accurate but also computationally efficient. Traditional methods, although effective, have faced challenges in scalability and dynamic monitoring. To address this gap, this paper introduces GeoViT, a novel vision transformer (ViT) based architecture designed specifically for geospatial GHG emission monitoring tasks. The introduction of GeoViT marks a significant stride in the realm of deep learning applied to environmental monitoring, setting a new benchmark for real-time, efficient GHG emission tracking and providing a robust foundation for future climate change mitigation efforts.

\section{Related Work}

Traditional models for spatial distribution of airborne pollutants like GHGs often rely on point measurements from specific locations, interpolated using geostatistical methods like kriging \cite{kriging} or Land-Use Regression (LUR) \cite{hoek2008}. While effective, these approaches require extensive variable selection and are limited in scalability. Recent advancements have been made in multitask deep learning for enhanced GHG quantification, where models simultaneously predict plume coverage, fuel type, and power generation rates from satellite images from Sentinel-2 to estimate CO\textsubscript{2} emissions \cite{multimodalCO2}. This approach significantly improved estimation accuracy, paving the way for better global emission monitoring. For NO\textsubscript{2} emission estimation, \cite{deepNO2} presents a deep learning model that leverages remote sensing data from Sentinel-2 and Sentinel-5P, allowing high-resolution and temporally informed estimates of surface-level NO\textsubscript{2} concentration. These methods overcome the limitations of static land-use datasets, demonstrating high accuracy against ground station measurements. Building on these works, our proposed model incorporates a ViT \cite{dosovitskiy2020}, outperforming previous models on key metrics while operating with reduced model sizes. Our work extends the capabilities of satellite-based environmental monitoring by leveraging the latest advances in ViT for improved GHG prediction tasks.

\section{Methodology}

\subsection{Mathematical and Algorithmic Advantages}

ViTs demonstrate a marked advantage in capturing long-range dependencies through their self-attention mechanism. This mechanism, as defined in \cite{dosovitskiy2020}, calculates the attention scores based on:
\begin{equation}
\text{Attention}(Q, K, V) = \text{softmax}\left(\frac{QK^T}{\sqrt{d_k}}\right)V
\end{equation}
This allows each element to attend over all positions, compared to convolutional neural networks (CNNs) that are limited by the size of their kernels \cite{wang2021pyramid}. In CNNs, the convolution operation is given by:
\begin{equation}
\text{Conv}(x) = \sum_{i=1}^{k}w_i \cdot x_{i}
\end{equation}
This fixed pattern of attention, dictated by the kernel size \( k \), restricts the CNN's ability to integrate information from distant spatial regions without deeper network architectures or larger convolution windows, which are computationally expensive \cite{guo2022transformer}.

Furthermore, ViTs offer enhanced parallel processing over RNNs for non-sequential data, which, along with positional embeddings, retains the sequence order crucial for spatial relation tasks \cite{touvron2020training}.

\subsection{Architecture}

\begin{figure}[htbp]
\centering
\includegraphics[width=\columnwidth]{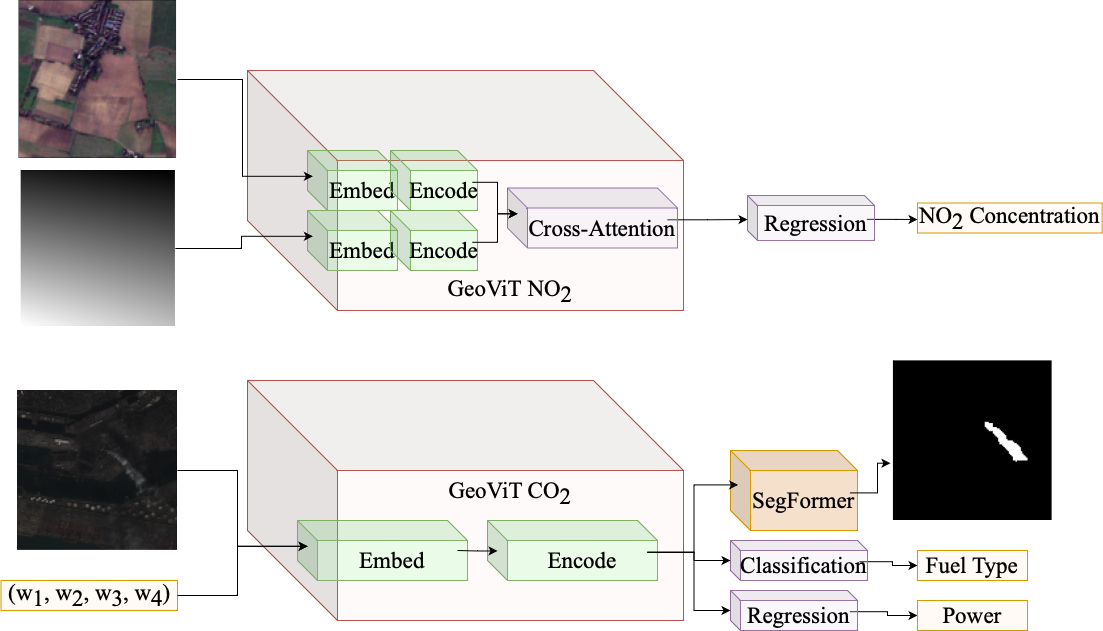}
\caption{Architecture of GeoViT for surface-level NO\textsubscript{2} concentration regression and multimodal analysis of CO\textsubscript{2} smoke plumes.}
\label{fig:architecture}
\end{figure}

Common to both CO\textsubscript{2} and NO\textsubscript{2} variants is the adoption of a ViT-based backbone. In both models, the initial stage of this backbone involves converting input images into a sequence of flattened patches. These patches are then linearly projected into an embedded space, where positional information is encoded to preserve the spatial context. The encoder, consisting of multi-headed self-attention \cite{voita2019analyzing} and feed-forward layers, processes these embeddings, which are then fed to task-specific heads. This architecture is able to capturing both local and global dependencies within the image. Both variants of GeoViT are depicted in Fig. \ref{fig:architecture}.

\subsubsection{CO\textsubscript{2} Variant}
The CO\textsubscript{2} variant takes as input a Sentinel-2 image and a weather vector a (temperature, relative humidity and wind speed) and produces a multimodal output that consists of a segmentation of smoke plums, classification of type of fired fuel, and estimated power generation rates, as described in \cite{multimodalCO2}. This variant integrates a specialized segmentation head based SegFormer architecture \cite{xie2021segformer}, which utilizes a lightweight MLP-based decoder that aggregates multi-scale features, providing both fine-grained local details and broader contextual information crucial for accurate segmentation. Additionally, the CO\textsubscript{2} variant includes dedicated heads consisting of fully-connected layers for classification and regression tasks.

\subsubsection{NO\textsubscript{2} Variant}
The NO\textsubscript{2} variant takes as input a Sentinel-2 image and a Sentinel-5P image, and employs a cross-attention mechanism between the two. This approach effectively learns to fuse features from Sentinel-2 and Sentinel-5P images \cite{lin2022cat}, and to encourage capturing of relevant interactions between the two input images. The output of the ViT backbone is fed to a regression head, which ouputs a scalar value for surface-level NO\textsubscript{2} concentration.

\subsection{Training and Implementation}

Both CO\textsubscript{2} and NO\textsubscript{2} variants of GeoViT were trained using the AdamW optimizer \cite{loshchilov2017decoupled}. Label smoothing \cite{szegedy2016rethinking} was incorporated as a regularization technique to improve model generalization. Data augmentation techniques such as random cropping, horizontal flipping, and brightness/contrast adjustments were employed to introduce variability in the training data, thus preventing overfitting and improving model resilience to variations in real-world satellite imagery.

\section{Experiments and Results}

\subsection{Experimental Setup}
We follow established protocols from previous literature on multi-modal CO\textsubscript{2} plume segmentation \cite{multimodalCO2} and surface-level NO\textsubscript{2} concentration regression \cite{deepNO2} to ensure consistency with existing benchmarks. Performance was evaluated using Segmentation IOU and Classification Accuracy for multi-modal CO\textsubscript{2} plume segmentation, and Top R2 Score, MAE, and MSE for surface-level NO\textsubscript{2} concentration regression.

\subsection{Results and Discussion}
We present a comparative analysis of the previous CNN-based models from \cite{multimodalCO2} and \cite{deepNO2}, and the proposed GeoViT architecture in table \ref{table:combined_table}. For the NO\textsubscript{2} estimation task, we do not pre-train on BigEarthNet \cite{sumbul2019bigearthnet} and therefore compare our model with the non pre-trained baseline in \cite{deepNO2}.

\begin{table}[h]
\caption{Comparison of CO\textsubscript{2} and NO\textsubscript{2} Task Performance}
\centering
\label{table:combined_table}
\begin{tabular}{|c|c|c|c|}
\hline
Task & Metric & Previous Model & GeoViT \\
\hline
CO\textsubscript{2} & Model Size (MB) & 3464.44 & 1374.07 \\
                     & Segmentation IOU & 0.668 & 0.724 \\
                     & Classification Accuracy & 0.853 & 0.99 \\
\hline
NO\textsubscript{2} & Model Size (MB) & 1964.46 & 850.58 \\
                     & Top R2 Score & 0.43 & 0.545 \\
                     & MAE & 6.68 & 5.847 \\
                     & MSE & 78.4 & 58.995 \\
\hline
\end{tabular}
\end{table}

We demonstrate that GeoViT's vision transformer backbone demonstrates superior performance over traditional CNN backbones for remote sensing images. Transformers excel in handling out-of-distribution samples, a common challenge in satellite imagery analysis \cite{wang2023cnns}. In the NO\textsubscript{2} task, the cross-attention mechanism between Sentinel-2 and Sentinel-5P images plays a crucial role in attending to relevant details between the two types of data. The resultant model is one that can significantly outperform CNN-based remote sensing models whilst also being much more computationally efficient.

\subsection{Conclusions}
The proposed GeoViT model demonstrated significant advancements over traditional CNN-based approaches. Our findings illustrate that GeoViT achieves superior accuracy in tasks like multimodal segmentation, classification, and regression for CO\textsubscript{2} and NO\textsubscript{2} emissions from satellite imagery. Key to this success is the model's ability to effectively capture long-range dependencies, resulting in more efficient and accurate environmental monitoring. Looking ahead, we aim to explore the incorporation of temporal patterns in remote sensing imagery over specific locations. This future work seeks to further refine our model's performance, enhancing its capability to monitor and predict environmental changes.

\bibliographystyle{IEEEtranS}
\bibliography{GeoViT_Paper}

\end{document}